\pdfoutput=1

\documentclass[11pt]{article}

\usepackage{acl}

\usepackage{times}
\usepackage{latexsym}

\usepackage[T1]{fontenc}

\usepackage[utf8]{inputenc}

\usepackage{microtype}
\usepackage{svg}
\usepackage{amsmath,amssymb}
\usepackage{here}
\usepackage{url}
\usepackage{comment}
\usepackage{booktabs}
\usepackage{multirow}
\usepackage{tabularx}
\usepackage{soul}
\usepackage{pifont}
\usepackage{tcolorbox}
\usepackage{CJK}
\usepackage{colortbl}
\usepackage{dsfont}

\newcommand{\ccg}{\cellcolor[gray]{0.9}}

\title{Look to the Right: Mitigating Relative Position Bias\\
in Extractive Question Answering}

\author{Kazutoshi Shinoda$^{1,2}$ ~~ Saku Sugawara$^2$ ~~ Akiko Aizawa$^{1,2}$\\
    $^1$The University of Tokyo\\
    $^2$National Institute of Informatics\\
    {\tt shinoda@is.s.u-tokyo.ac.jp} \\
    {\tt \{saku,aizawa\}@nii.ac.jp} \\
}

\begin{document}
\maketitle
\begin{abstract}
Extractive question answering (QA) models tend to exploit spurious correlations to make predictions when a training set has unintended biases.
This tendency results in models not being generalizable to examples where the correlations do not hold.
Determining the spurious correlations QA models can exploit is crucial in building generalizable QA models in real-world applications; moreover, a method needs to be developed that prevents these models from learning the spurious correlations even when a training set is biased.
In this study, we discovered that the relative position of an answer, which is defined as the relative distance from an answer span to the closest question-context overlap word, can be exploited by QA models as superficial cues for making predictions.
Specifically, we find that when the relative positions in a training set are biased, the performance on examples with relative positions unseen during training is significantly degraded.
To mitigate the performance degradation for unseen relative positions, we propose an ensemble-based debiasing method that does not require prior knowledge about the distribution of relative positions.
We demonstrate that the proposed method mitigates the models' reliance on relative positions using the biased and full SQuAD dataset.
We hope that this study can help enhance the generalization ability of QA models in real-world applications.\footnote{Our codes are available at \url{https://github.com/KazutoshiShinoda/RelativePositionBias}.}
\end{abstract}

\section{Introduction}
Deep learning-based natural language understanding (NLU) models are prone to use spurious correlations in the training set.
This tendency results in models' poor generalization ability to out-of-distribution test sets \cite{mccoy-etal-2019-right,geirhos_shortcut_2020}, which is a significant challenge in the field.
Question answering (QA) models trained on intentionally biased training sets are more likely to learn solutions based on spurious correlations rather than on causal relationships between inputs and labels.
For example, QA models can learn question-answer type matching heuristics \cite{lewis-and-fan-2019-generative}, and absolute-positional correlations \cite{ko-etal-2020-look}, particularly when a training set is biased toward examples with corresponding spurious correlations.
Collecting a fully unbiased dataset is challenging.
Therefore, it is vital to discover possible dataset biases that can degrade the generalization and develop debiasing methods to learn generalizable solutions even when training on unintentionally biased datasets.

\begin{table}[t]
    \centering
    \small
    \begin{tabular}{p{1.1cm}p{5.5cm}}
        \toprule
        Context & ... This changed \ul{in} \textbf{1924} with formal requirements developed for graduate degrees, including offering \ul{Doctorate} (PhD) \ul{degrees} ... \\
        \midrule
        Question & The granting of \ul{Doctorate} \ul{degrees} first occurred \ul{in} what year at Notre Dame? \\
        \midrule
        Relative Position & $-1$ \\
        \midrule\midrule
        Context & ... \ul{The} other magazine, \ul{The} \ul{Juggler}, \ul{is} released \textbf{twice} a year and focuses on student literature and artwork ... \\\midrule
        Question & How often \ul{is} Notre Dame's \ul{the} \ul{Juggler} published? \\\midrule
        Relative Position & $-2$ \\
        \bottomrule
    \end{tabular}
    \caption{Examples taken from SQuAD. \ul{Underlined} words are contained in both the context and question. \textbf{Bold} spans are the answers to the questions. In both the examples, answers are found by \emph{looking to the right} from the overlapping words. See \S\ref{sec:definition} for the definition of the relative position.}
    \label{tb:example}
\end{table}

\begin{figure}[t]
\centering
\includegraphics[clip,width=7.9cm]{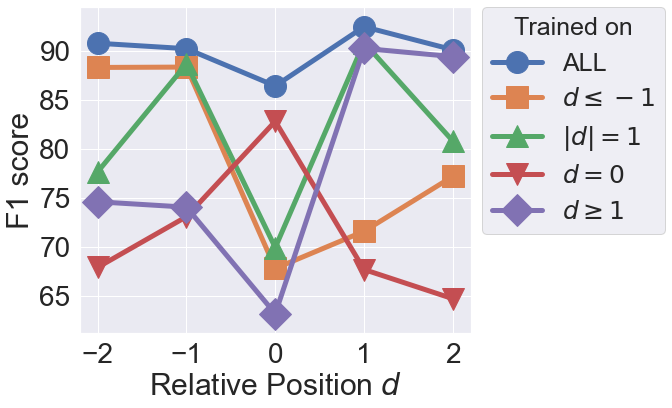}
\caption{F1 score for each relative position $d$ in the SQuAD development set. ``ALL'' in the legend refers to a QA model trained on all the examples in the SQuAD training set. The other terms refer to models trained only on examples for which the respective conditionals are satisfied.
BERT-base was used for the QA models.
The accuracy is comparable to ALL for examples with seen relative positions, but worse for others.
Please refer to \S\ref{sec:definition} for the definition of $d$.}
\label{fig:concept}
\end{figure}

In extractive QA \cite[e.g.,][]{rajpurkar-etal-2016-squad}, in which answers to questions are spans in textual contexts, we find that the relative position of an answer, which is defined as the relative distance from an answer span to the closest word that appears in both a context and a question, can be exploited as superficial cues by QA models.
See Table \ref{tb:example} for the examples.
Specifically, we find that when the relative positions are intentionally biased in a training set, a QA model tends to degrade the performance on examples where answers are located in relative positions unseen during training, as shown in Figure \ref{fig:concept}.
For example, when a QA model is trained on examples with negative relative positions, as shown in Table \ref{tb:example}, the QA performance on examples with non-negative relative positions is degraded by 10 $\sim$ 20 points, as indicated by the square markers ($\blacksquare$) in Figure \ref{fig:concept}.
Similar phenomena were observed when the distribution of the relative positions in the training set was biased differently, as shown in Figure \ref{fig:concept}.
This observation implies that the model may preferentially learn to find answers from seen relative positions.

We aim to develop a method for mitigating the performance degradation on subsets with unseen relative positions while maintaining the scores on subsets with seen relative positions, even when the training set is biased with respect to relative positions.
To this end, we propose debiasing methods based on an ensemble \cite{hinton-2002-training} of intentionally biased and main models.
The biased model makes predictions relying on relative positions, which promotes the main model not depending solely on relative positions.
Our experiments on SQuAD \cite{rajpurkar-etal-2016-squad} using BERT-base \cite{devlin-etal-2019-bert} as the main model show that the proposed methods improved the scores for unseen relative positions by 0$\sim$10 points.
We demonstrate that the proposed method is effective in four settings where the training set is differently filtered to be biased with respect to relative positions.
Furthermore, when applied to the full training set, our method improves the generalization to examples where questions and contexts have no lexical overlap.

\section{Relative Position Bias}
\subsection{Definition}
\label{sec:definition}
In this study, we call a word that is contained in both the question and the context as an overlapping word.
Let $d$ be the relative position of the nearest overlapping word from the answer span in extractive QA.
If $w$ is a word, $c=\{w_i^c\}_{i=0}^N$ for the sentence, $q=\{w_i^q\}_{i=0}^M$ for the question, and $a=\{w_i^c\}_{i=s}^{e}$ ($0 \leq s \leq e \leq N$) for the answer, the relative position $d$ is defined as follows:
\begin{align}
& f(j, s, e) = 
    \begin{cases}
    j - s,  & \text{for }j < s\\
    0,      & \text{for }s \leq j \leq e\\
    j - e,  & \text{for }j > e
    \end{cases}\\
& D = \{ f(j, s, e) | w_j^c \in q \} \label{eq:D}\\
& d = \mathrm{argmin}_{d' \in D} |d'| \label{eq:d}
\end{align}
where $0 \leq j \leq N$ denotes the position of the word $w_j^c$ in the sentence, $f(i, s, e)$ denotes the relative position of $w_i^c$ from $a$, and $D$ denotes the set of relative positions of all overlapping words.\footnote{Because function words as well as content words are important clues for reading comprehension, $D$ in Equation \ref{eq:D} can contain function and content words.}
Because QA models favor spans that are located close to the overlapping words \cite{jia-liang-2017-adversarial} and accuracy deteriorates when the absolute distance between the answer span and the overlapping word is considerable \cite{sugawara-etal-2018-makes}, the one with the lowest absolute value in Equation \ref{eq:d} is used as the relative position.\footnote{There are a few cases where $d$ in Equation \ref{eq:d} is not fixed to one value. However, such examples are excluded from the training and evaluation sets for brevity.}

\subsection{Distribution of Relative Position $d$}
\label{sec:rel-pos-dist}
\begin{figure}[t]
\centering
\includegraphics[clip,width=6cm]{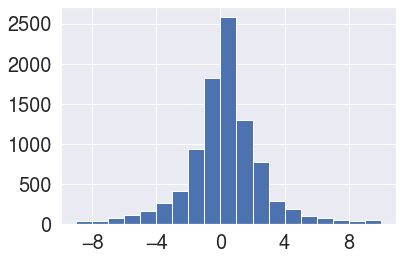}
\caption{Histogram of relative position $d$ in the SQuAD training set.}
\label{fig:hist}
\end{figure}

Figure \ref{fig:hist} shows the distribution of relative position $d$ in the SQuAD \cite{rajpurkar-etal-2016-squad} training set.
This demonstrates that the $d$ values are biased around zero.
Although the tendency to bias around zero is consistent for the other QA datasets, there are differences in the distribution between the datasets.
See Appendix \ref{app:hist} for more details.
This difference may be caused by how the datasets were collected or to the contexts' domains.
Therefore, building a QA model that does not overfit a specific distribution of relative positions is necessary.

\begin{figure*}[tbp]
\centering
\includegraphics[width=14.5cm]{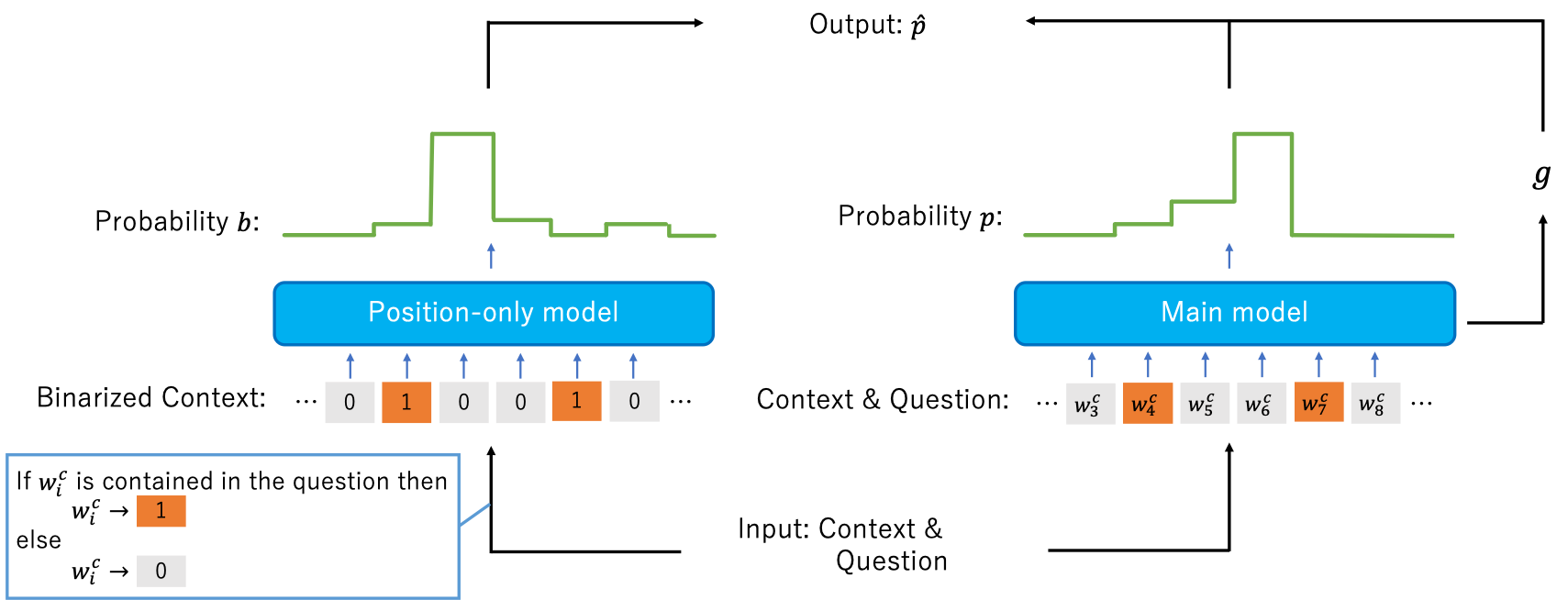}
\caption{Illustration of LearnedMixin with Position-only model (PosOnly) as a biased model. The output $\hat{p}$ is used for computing the loss, and the probability $p$ is used for inference.}
\label{fig:overview}
\end{figure*}

\section{Method}
\subsection{Debiasing Algorithm}
\label{sec:debias}
For debiasing algorithms, we employ BiasProduct and LearnedMixin \cite{clark-etal-2019-dont,he-etal-2019-unlearn}, which are based on product-of-experts \cite{hinton-2002-training}, following \citet{ko-etal-2020-look}.
In both methods, after an intentionally biased model is prepared, the cross-entropy loss is calculated using a product of the biased and main models.
The biased model is fixed when minimizing the loss to train the main model.
Only the main model is used to make predictions during testing.
Following existing research \cite{seo2017bidirectional,devlin-etal-2019-bert}, the model $\hat{p}$ outputs the probabilities $\hat{p}(s)$ and $\hat{p}(e)$ of the start position $s$ and the end position $e$ of the answer span, and the loss function is the sum of the cross-entropy of the start and end positions.
For simplicity, $\hat{p}(s)$ and $\hat{p}(e)$ will be denoted as $\hat{p}$.

\subsubsection{BiasProduct}
In BiasProduct, the sum of the logarithms of the output probability $b$ of the biased model and the output probability $p$ of the main model is given to the softmax function to obtain $\hat{p}$, as follows.
\begin{align}
\hat{p} = \textrm{softmax}(\log p + \log b)
\end{align}
This encourages the the main model to learn examples where the biased model make incorrect predictions, rather than examples that contain biases allowing the biased model to make correct predictions.

\subsubsection{LearnedMixin}
BiasProduct strongly depends on the output probability of the biased model.
The main model can be made more robust by using LearnedMixin, which predicts whether the main model can trust the prediction of the biased model for each example.
\begin{align}
\hat{p} = \textrm{softmax}(\log p + g(c, q)\log b)
\end{align}
where $g (\geq 0)$ is a learnable function that takes $q$ and $c$ as inputs.

\subsection{Biased Model}
We describe how to construct the biased model described in \S\ref{sec:debias}.
The first model, Answer Prior, is a fully rule-based model with a prior probability of answer span.
The second, the Position-only model, is a QA model trained with only binarized contexts from which models can know only which token is an overlapping word.

\subsubsection{Answer Prior (AnsPrior)}
We first use a simple heuristic called AnsPrior as a biased model.
AnsPrior empirically defines the prior probabilities of the start and end positions of the answer span $a$ according to the distribution of the relative position $d$ in a training set.
The prior probability $b_i$ that a word $w_i^c$ in a sentence is a start or end of the answer is defined as follows for each of the subsets of the training set that satisfies one of the four conditions shown in the legend of Figure~\ref{fig:concept}.
\begin{align}
& b_i =\nonumber\\
&
\begin{cases}
\mathds{1}\left[ w_{i+1}^c \in q  \right]/Z, & \text{for }d \leq -1\\
\mathds{1}\left[ (w_{i+1}^c \in q) \lor (w_{i-1}^c \in q)  \right]/Z, & \text{for }|d| = 1\\
\mathds{1}\left[ w_i^c \in q \right]/Z, & \text{for }d = 0\\
\mathds{1}\left[ w_{i-1}^c \in q \right]/Z, & \text{for }d \geq 1
\end{cases}
\end{align}
where $\mathds{1}$ denotes an indicator function (e.g., $\mathds{1}\left[ w_{i+1}^c \in q  \right]$ returns 1 if $w_{i+1}^c$ is contained in $q$, otherwise it returns 0) and $Z$ denotes a normalizing constant.
These prior probabilities are based on a heuristic that assigns equal probabilities to the possible answers in the training set.
Therefore, they are inflexible as they are prior probabilities specific to the distribution of relative positions of a training set.

\subsubsection{Position-only model (PosOnly)}
We propose PosOnly as a biased model that can be used without prior knowledge about the distributions of relative positions in training sets.
PosOnly accepts as input the sequence of binary variables indicating if a word in a context is overlapped with a question.
The only information available to PosOnly to predict answer spans is the relative distances from the overlapping words.
Hence, it is expected to learn solutions using the relative positions regardless of how biased the relative positions of the training set are.
The illustration of LearnedMixin with PosOnly is shown in Figure \ref{fig:overview}.

\section{Experiments}

\subsection{Generalization to Unseen Relative Positions}
\label{sec:main-experiment}

\paragraph{Dataset}
SQuAD 1.1 \cite{rajpurkar-etal-2016-squad} was used as the dataset.
The training set is filtered to be biased in four different ways to assess the applicability of our methods.
The four subsets were constructed by extracting only examples whose relative positions $d$ satisfied the conditions $d \leq -1$, $|d| = 1$, $d=0$, and $d \geq 1$.
The sizes of the subsets are 33,256, 30,003, 21,266, and 25,191, respectively.
The scores of BERT-base trained on the full training set are also reported for comparison.
For evaluation, we reported the F1 scores on subsets of the SQuAD development set stratified by the relative positions.

\begin{table*}[!t]
    \centering
    \begin{tabular}{ccccccccc}
\toprule
 & & \multicolumn{7}{c}{Evaluated on} \\
\cmidrule{3-9}
Trained on & Model & \begin{tabular}{c}
  $d \leq$\\
  $-3$
\end{tabular} & \begin{tabular}{c}
  $d =$\\
  $-2$
\end{tabular} & \begin{tabular}{c}
  $d =$\\
  $-1$
\end{tabular} & $d = 0$ & $d = 1$ & $d = 2$ & $d \geq 3$\\
\midrule
ALL & BERT-base & \ccg 82.19 & \ccg 90.82 & \ccg 90.25 & \ccg 86.47 & \ccg 92.49 & \ccg 90.14 & \ccg 81.43\\
\midrule
$d \leq -1$ & BERT-base & 78.17 \ccg & 88.34 \ccg & 88.38 \ccg & 67.82 & 71.62 & 77.22 & 69.54\\
$d \leq -1$ & BiasProduct-AnsPrior & 73.00 \ccg & 84.34 \ccg & 85.61 \ccg & 46.32 & 25.23 & 64.91 & 59.06\\
$d \leq -1$ & LearnedMixin-AnsPrior & 79.07 \ccg & 89.27 \ccg & 89.01 \ccg & 68.52 & 72.35 & 80.43 & 70.31\\
$d \leq -1$ & BiasProduct-PosOnly & 75.04 \ccg & 83.90 \ccg & 83.22 \ccg & 73.80 & 81.35 & 81.79 & 73.27\\
$d \leq -1$ & LearnedMixin-PosOnly & 77.00 \ccg & 86.72 \ccg & 86.25 \ccg & 74.26 & 82.66 & 82.81 & 75.94\\
\midrule
$|d| = 1$ & BERT-base & 65.62 & 77.69 & 88.70 \ccg & 69.96 & 90.88 \ccg & 80.84 & 66.42\\
$|d| = 1$  & BiasProduct-AnsPrior & 60.44 & 75.07 & 56.44 \ccg & 49.32 & 52.37 \ccg & 72.85 & 57.98\\
$|d| = 1$  & LearnedMixin-AnsPrior & 73.42 & 83.39 & 88.70 \ccg & 74.24 & 90.47 \ccg & 85.51 & 73.52\\
$|d| = 1$ & BiasProduct-PosOnly & 72.41 & 80.59 & 84.01 \ccg & 73.34 & 87.61 \ccg & 83.11 & 72.09\\
$|d| = 1$ & LearnedMixin-PosOnly & 73.76 & 80.63 & 86.10 \ccg & 74.50 & 89.64 \ccg & 82.98 & 72.04\\
\midrule
$d = 0$ & BERT-base & 60.75 & 67.94 & 73.11 & 82.85 \ccg & 67.72 & 64.74 & 52.88\\
$d = 0$ & BiasProduct-AnsPrior & 56.25 & 65.15 & 69.05 & 81.07 \ccg & 65.10 & 62.95 & 49.43\\
$d = 0$ & LearnedMixin-AnsPrior & 59.66 & 69.62 & 72.53 & 83.06 \ccg & 68.04 & 66.03 & 53.29\\
$d = 0$ & BiasProduct-PosOnly & 62.97 & 67.88 & 70.22 & 78.66 \ccg & 66.69 & 69.12 & 59.88\\
$d = 0$ & LearnedMixin-PosOnly & 65.09 & 70.47 & 72.51 & 81.32 \ccg & 68.29 & 68.47 & 59.54\\
\midrule
$d \geq 1$ & BERT-base & 68.03 & 74.63 & 74.08 & 63.21 & 90.28 \ccg & 89.44 \ccg & 75.42 \ccg\\
$d \geq 1$ & BiasProduct-AnsPrior & 58.63 & 63.13 & 29.08 & 39.22 & 88.53 \ccg & 88.34 \ccg & 72.29 \ccg\\
$d \geq 1$ & LearnedMixin-AnsPrior & 70.71 & 77.22 & 76.82 & 66.67 & 90.87 \ccg & 89.75 \ccg & 76.31 \ccg\\
$d \geq 1$ & BiasProduct-PosOnly & 68.54 & 78.13 & 78.58 & 70.72 & 85.17 \ccg & 81.59 \ccg & 72.90 \ccg\\
$d \geq 1$ & LearnedMixin-PosOnly & 71.17 & 80.41 & 79.97 & 71.33 & 87.53 \ccg & 84.33 \ccg & 74.24 \ccg\\
\bottomrule
    \end{tabular}
    \caption{F1 scores for each subset of the SQuAD development set.
    The cells with relative position $d$ seen during training are indicated by \begin{tabular}{c}gray\ccg\end{tabular}.
    For \begin{tabular}{c}gray\ccg\end{tabular} cells, the scores tend to remain close to those in the case where the full training set is used (ALL).
    Conversely, the scores for the other white cells tend to be lower than ALL.
    }
    \label{tb:main-result}
\end{table*}

\paragraph{Method}
We compare four combinations of two learning methods for debiasing, BiasProduct and LearnedMixin, and two biased models, AnsPrior and PosOnly.
BERT-base \cite{devlin-etal-2019-bert} was used for both the main model and PosOnly.
We also evaluate the BERT-base with standard training as the baseline.
The training details are given in Appendix \ref{app:training}.

\paragraph{Results} Table \ref{tb:main-result} shows the results.
First, when the full training set is used (ALL), the performance of the BERT-base baseline exceeds 90 points when $|d| = 1, 2$, whereas it drops by about eight points when $|d| \geq 3$.
As shown in \S\ref{sec:rel-pos-dist}, there is a correlation between the accuracy for a specific range of relative positions and the frequency of examples whose relative positions are in the corresponding range in the training set.

The results of the BERT-base baseline trained on subsets where the distribution of relative position $d$ was intentionally biased strengthened the credibility of this hypothesis.
For example, when the standard training was performed only on examples with relative positions $|d|=1$, the F1 score decreased by less than two points for $|d|=1$ compared to ALL, whereas the F1 score decreased by 10$\sim$15 points for $|d| \neq 1$.
A similar trend was observed in the BERT-base baseline trained on other subsets.
This suggests that the model exploited spurious correlations regarding relative positions in the biased subsets to make predictions.

We compare the four proposed debiasing methods under the same conditions of relative positions in the training sets.
For the debiasing algorithms, LearnedMixin produced higher F1 scores than BiasProduct in most cases.
This result shows the effectiveness of learning the degree to which the predictions of a biased model should be utilized for training the main model.
For the biased models, PosOnly was superior to AnsPrior for improving the generalization ability to examples with relative positions unseen during training, i.e., the scores in white cells in Table \ref{tb:main-result}.
LearnedMixin-PosOnly outperformed LearnedMixin-AnsPrior by about 5 points when trained on $d=0$ and tested on $d \leq -3$, and when trained on $d \leq -1$ and tested on $d \geq 3$.

In contrast, regarding the scores on examples with relative positions seen during training (i.e., the scores in the \begin{tabular}{c}\ccg gray\end{tabular} cells in Table \ref{tb:main-result}), LearnedMixin-AnsPrior was superior to LearnedMixin-PosOnly.
As pointed out in \citet{utama-etal-2020-mind}, the trade-off between accuracies on in- and out-of-distribution test sets was observed in our cases.
Mitigating the trade-off for relative positions is future work.

\begin{table*}[th]
    \centering
    \begin{tabular}{cccc}
\toprule
 & & \multicolumn{2}{c}{Evaluated on} \\
\cmidrule{3-4}
Trained on & Model & $c \cap q\neq\phi$ & $c \cap q=\phi$\\
\midrule
ALL & BERT-base & 87.94 & 67.11\\
ALL & BiasProduct-PosOnly & 84.83 & 59.88\\
ALL & LearnedMixin-PosOnly & 87.37 & 80.44\\
\bottomrule
    \end{tabular}
    \caption{F1 scores for two subsets of the SQuAD development set.
    Each model is trained on the full SQuAD training set.
    $c$ indicates the context, and $q$ indicates the question.
    $\phi$ indicates the empty set.
    The scores for each relative position are given in Appendix \ref{app:detail}.
    }
    \label{tb:full-training-set}
\end{table*}

\subsection{Effect of Mitigating Relative Position Bias in Normal Settings}
\label{sec:effect}
Although we verified the effectiveness of our methods on intentionally biased datasets in \S\ref{sec:main-experiment}, the proposed biased model, PosOnly, can also be applied to training on standard datasets because it does not require prior information about the distribution of relative positions, unlike AnsPrior.
To investigate the effect of our method in a normal setting, we first trained PosOnly on the full training set.
We then trained BERT-base as the main model on the full training set using BiasProduct or LearnedMixin with PosOnly as the biased model.

The results are shown in Table \ref{tb:full-training-set}.
LearnedMixin-PosOnly improved the generalization to a subset where questions and contexts have no common words (i.e., $c \cap q=\phi$), with little performance degradation on the other subset (i.e., $c \cap q \neq \phi$).
This observation implies that our method might mitigate the reliance on overlapping words in a normal setting.
However, the size of the subset $c \cap q=\phi$ is only 15, which makes the above conclusion unreliable.
Future work should increase the size of this subset and verify its effectiveness.



\section{Related Work}
Recent studies have found that NLU models tend to learn shortcut solutions specific to the distribution of the training sets.
In natural language inference, models can use spurious correlations regarding lexical items \cite{gururangan-etal-2018-annotation} and lexical overlap \cite{mccoy-etal-2019-right}.
In extractive QA, existing studies have shown that models can learn several kinds of shortcut solutions.
\citet{weissenborn-etal-2017-making} showed that a substantial number of questions could be answered only by matching the types of questions and answers.
\citet{sugawara-etal-2018-makes} demonstrated that only using partial inputs is sufficient for finding correct answers in most cases.
\citet{ko-etal-2020-look} indicated that the absolute position bias of answers could severely degrade the generalization.
\citet{shinoda-etal-2021-question} showed that question generation models can amplify lexical overlap bias when used for data augmentation in QA.

Sampling training and test sets from different distributions \cite{jia-liang-2017-adversarial,fisch-etal-2019-mrqa,lewis-and-fan-2019-generative,ko-etal-2020-look} is one of the most effective frameworks to assess whether models can learn generalizable solutions.
In this study, we also employed the framework to analyze the generalization capability from a new perspective of relative position bias, and developed effective debiasing methods for this bias.
Specifically, we used the same loss functions as \citet{ko-etal-2020-look}, and proposed new biased models in this study.

\section{Conclusion}
We showed that an extractive QA model tends to exploit relative position bias in training sets, causing the performance degradation for relative positions unseen during training.
To mitigate this problem, we proposed new biased models that perform well with existing debiasing algorithms on intentionally filtered training sets.
Furthermore, when applied to the full training set, our method improved the generalization to examples where questions and contexts have no common words.
Future work includes refining the definition of relative position bias that is more learnable for QA models, mitigating the trade-off between the accuracies for seen and unseen relative positions, and increasing the size of the test set with no lexical overlap between question and context.

\section*{Acknowledgements}
We would like to thank the anonymous reviewers for their valuable comments.
This work was supported by JST SPRING Grant Number JPMJSP2108 and JSPS KAKENHI Grant Numbers 22J13751 and 22K17954.
This work was also supported by the SIP-2 ``Big-data and AI-enabled Cyberspace Technologies'' by the New Energy and Industrial Technology Development Organization (NEDO).

\bibliography{anthology,custom}

\appendix

\begin{figure*}[htbp]
    \centering
    \begin{tabular}{cc}
         \includegraphics[clip,width=7cm]{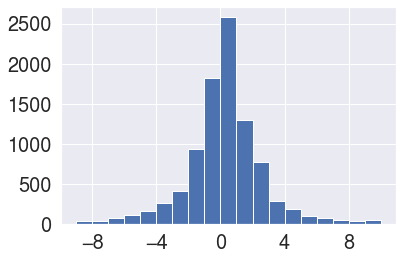} & \includegraphics[clip,width=7cm]{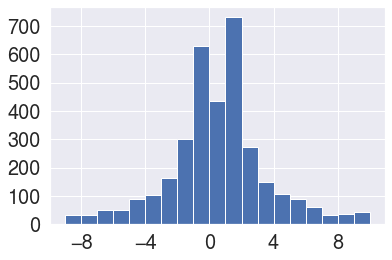}\\
         (a) SQuAD & 
         (b) NewsQA\\
         \includegraphics[clip,width=7cm]{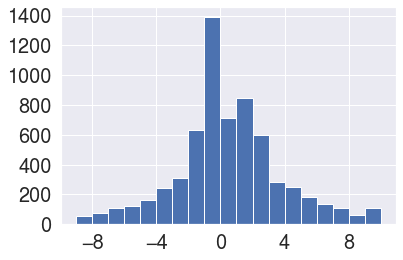} & \includegraphics[clip,width=7cm]{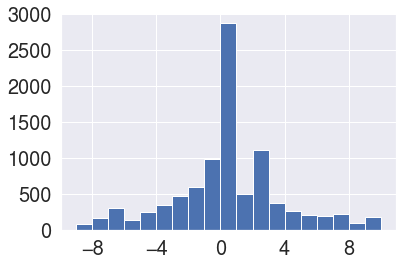} \\
         (c) TriviaQA & (d) NaturalQuestions\\
    \end{tabular}
    \caption{Histograms of relative position $d$ in the SQuAD, NewsQA, TriviaQA, and NaturalQuestions development sets.}
    \label{fig:other-datasets}
\end{figure*}

\begin{table*}[!t]
    \centering
    \begin{tabular}{ccccccccc}
\toprule
 & & \multicolumn{7}{c}{Evaluated on} \\
\cmidrule{3-9}
Trained on & Model & \begin{tabular}{c}
  $d \leq$\\
  $-3$
\end{tabular} & \begin{tabular}{c}
  $d =$\\
  $-2$
\end{tabular} & \begin{tabular}{c}
  $d =$\\
  $-1$
\end{tabular} & $d = 0$ & $d = 1$ & $d = 2$ & $d \geq 3$\\
\midrule
ALL & BERT-base &  82.19 &  90.82 &  90.25 &  86.47 &  92.49 &  90.14 &  81.43\\
ALL & BiasProduct-PosOnly & 81.04  & 85.72  & 86.89  & 83.53  & 88.47  & 87.16  & 80.61 \\
ALL & LearnedMixin-PosOnly & 82.48  & 90.63  & 90.15  & 85.69  & 91.33  & 89.14  & 81.21 \\
\bottomrule
    \end{tabular}
    \caption{F1 scores for each subset of the SQuAD development set.
    }
    \label{tb:detail}
\end{table*}

\section{Training Details}
\label{app:training}
The number of training epochs for each model was 2, the batch size was 32, the learning rate decreased linearly from 3e-5 to 0, and Adam \cite{kingma2014adam} was used for optimization.

\section{Distribution of Relative Position $d$}
\label{app:hist}
The frequency distribution of relative positions in the SQuAD, NewsQA, TriviaQA, and NaturalQuestions development sets is shown in Figure \ref{fig:other-datasets}.
Although there are differences among the datasets, they all show that the relative positions are biased around 0.

\section{Results on the Full Training Set}
\label{app:detail}
The detailed scores of our methods when trained on the full training set are shown in Table~\ref{tb:detail}.
Compared to the BERT-base baseline, LearnedMixin-PosOnly can maintain the score in each column.

\end{document}